\documentclass[]{article}
\usepackage{amssymb}
\usepackage{amsmath}
\usepackage{amsthm}

\usepackage{graphicx}

\title{Correlation of Data Reconstruction Error and Shrinkages in Pair-wise Distances under Principal 
    Component Analysis (PCA)}

\author{Abdulrahman Oladipupo Ibraheem \\ rahmanoladi@yahoo.com \\ \\ Computing and Intelligent Systems Research Group  \\ Department of Computer Science and Engineering \\ Obafemi Awolowo University, Ile-Ife, Nigeria. }

\begin{document}

\maketitle

\begin{abstract}
In this \lq on-going\rq $\:$ work, I explore certain theoretical and empirical implications of data transformations under the PCA.  In particular, I state and prove three theorems about PCA, which I paraphrase as follows:  1). PCA without discarding eigenvector rows is injective, but looses this injectivity when eigenvector rows are discarded  2). PCA without discarding eigenvector rows preserves pair-wise distances, but tends to cause pair-wise distances to shrink when eigenvector rows are discarded. 3). For any pair of points, the shrinkage in pair-wise distance is bounded above by an $L_1$ norm reconstruction error associated with the points. Clearly, “3).” suggests that there might exist  some correlation  between shrinkages in pair-wise distances and mean square reconstruction error which is defined as the sum of those eigenvalues associated with the discarded eigenvectors. I therefore decided to perform numerical experiments to obtain the correlation between the sum of those eigenvalues and shrinkages in pair-wise distances. In addition, I have also performed some experiments to check respectively the effect of the sum of those eigenvalues and the  effect of the shrinkages on classification accuracies under the PCA map. So far, I have obtained the following results on some publicly available data from the UCI Machine Learning Repository:  1). There seems to be a strong correlation between the sum of those eigenvalues associated with discarded eigenvectors and shrinkages in pair-wise distances. 2). Neither the sum of those eigenvalues nor pair-wise distances have any strong correlations with classification accuracies. 
\end{abstract}

\section{Introduction: A review of PCA}

In this introductory section, I will review the basics of PCA. A good reference on this topic is in \cite{cast95} Essentially, PCA involves two key aspects: transformation of data to a zero-correlation space, and truncation of the data. Depending on the application, one may choose not to truncate the data after the transformation. To proceed, let us denote the data to be transformed by \textbf{column} vectors $\textbf{x}_i$. The covariance matrix of the $\textbf{x}_i$'s can be written as:

\begin{equation}
  C_x = \frac{1}{N} \sum_{i=1} ^{N} (\textbf{x}_i - \overline{\textbf{x}})(\textbf{x}_i - \overline{\textbf{x}})^t
\end{equation}
     in which $\overline{\textbf{x}}$ is the mean of the $\textbf{x}_i$'s. Next, we consider the transformation
     \begin{equation}
     \textbf{y}_i = Q^t(\textbf{x}_i - \overline{\textbf{x}}) 
     \end{equation}

    \noindent where $Q^t$ is a $n$ by $n$ matrix whose $j$-th row, $(q_j)^t$, is the $j$-th eigenvector of the $n$ by $n$ symmetric matrix $C_x$. We shall see that the $\textbf{y}_i$'s have zero correlations (but possibly non-zero variances). The covariance matrix of the $\textbf{y}_i$'s may be expressed as:
               
       \begin{equation}
         C_y = \frac{1}{N} \sum_{i=1} ^{N} (\textbf{y}_i - \overline{\textbf{y}})(\textbf{y}_i - \overline{\textbf{y}})^t
       \end{equation}
            
       \noindent But the mean, $\overline{\textbf{y}}$, of the $\textbf{y}_i$'s is zero because:
               
         \begin{equation}
           \overline{\textbf{y}} = \frac{1}{N} \sum_{i=1} ^{N} \textbf{y}_i 
          \end{equation}
                
           \begin{equation}
                    \overline{\textbf{y}} = \frac{1}{N} \sum_{i=1} ^{N} Q^t(\textbf{x}_i - \overline{\textbf{x}}) 
            \end{equation}
                
                \begin{equation}
                     \overline{\textbf{y}} = \frac{1}{N} Q^t \; ( \sum_{i=1} ^{N}\textbf{x}_i - \sum_{i=1} ^{N}\overline{\textbf{x}}\,) 
                \end{equation}
                
                 \begin{equation}
                           \overline{\textbf{y}} = \frac{1}{N} Q^t \; ( N \overline{\textbf{x}} - N \overline{\textbf{x}}\,) \; = \; 0
                 \end{equation}
                
               \noindent To show that the $\textbf{y}_i$'s have zero correlations, we need to show that $C_y$ is diagonal. Putting Equation 3 into Equation 5, we find:
               
              \begin{equation}
                C_y = \frac{1}{N} \sum_{i=1} ^{N} \textbf{y}_i \textbf{y}_i^t             
              \end{equation} 
              
           Putting $\textbf{y}_i$ from Equation 2 into the preceding equation yields:

               \begin{equation}
                 C_y = \frac{1}{N} \sum_{i=1} ^{N} Q^t(\textbf{x}_i-\overline{\textbf{x}})\; (Q^t(\textbf{x}_i-\overline{\textbf{x}}))^t             
               \end{equation} 
           
           \begin{equation}
                  C_y = \frac{1}{N} \sum_{i=1} ^{N} Q^t(\textbf{x}_i-\overline{\textbf{x}})\; (\textbf{x}_i-\overline{\textbf{x}})^tQ             
           \end{equation}
            
          Above, we recognize $(\textbf{x}_i-\overline{\textbf{x}})\; (\textbf{x}_i-\overline{\textbf{x}})^t$ as $C_x$ and write:
          
           \begin{equation}
                  C_y = \frac{1}{N} \sum_{i=1} ^{N} Q^tC_xQ             
           \end{equation} 
            
            \noindent The preceding equation, when viewed in the context of the spectral theorem of linear algebra, informs us that $C_y$ must be diagonal. This is because $C_x$ is real symmetric, and therefore orthogonally diagonalizable; and because the rows of $Q^t$ are the eigenvectors of $C_x$. Moreover, according to another theorem of linear algebra, the principal-diagonal element $(C_y)_{jj}$, which is on the $j$-th row and $j$-th column of $C_y$ is the eigenvalue corresponding to the eigenvector lying in the $j$-th row of $Q^t$. \\ \\
             
             We have seen how PCA projects data to zero-correlation space. We will now describe how to harness it for dimensionality reduction. Suppose we wish to transform a $n$-dimensional vector, $\textbf{x}_i$  from $\mathbb{R}^{n}$ to $\mathbb{R}^{m}$, under the stipulation that $m < n$, so that the transformation yields a $m$-dimensional vector $\hat{\textbf{y}}_i$. Mathematically, the required transformation is summarized as follows:
             
              \begin{equation}
                    \hat{\textbf{y}} = \hat{Q}_t(\textbf{x} - \overline{\textbf{x}})
              \end{equation}
              
              Where $\hat{Q}_t$ is the $m$ by $n$ matrix obtained by discarding the lowest $n$ minus $m$ rows of ${Q}_t$. Observe that this is tantamount to discarding the $n - m$ eigenvectors of matrix $C_x$ that are associated with the $n - m$ lowest eigenvalues of ${Q}_t$.           
              
              It is clear that the above transformation truncates the vector $x_i$ in the sense that it causes it to lose $n - m$ of its dimensions. This truncation results in a root mean square error, $\hat{e}$, given by:
              
              \begin{equation}
                  \hat{e} \; = \; \sum_{i=n+1}^{m} \lambda_i
              \end{equation}

          where $\lambda_i$ is the eigenvalue associated with the $i$-th eigenvector, from top down, in the original $n$ by $n$ matrix $Q^t$. From the above equation, it is obvious that $\hat{e}$ is just the sum of the eigenvalues associated with the discarded eigenvectors. Thus, we see that error $\hat{e}$ is proportional to the number of discarded eigenvectors. 
          
          \section{Non-truncated versus truncated PCA transformations: two theorems and their proofs }
        The PCA transformation can either be ``truncated" or ``non-truncated," the former case occurring when one or more of the (eigenvector) rows of the transformation matrix are discarded, and the latter case occurring when no row is discarded.  In this section, using the background presented in the previous section, I shall state and prove two theorems which juxtapose the structures of the data when PCA is performed with and without truncation. The first theorem states that PCA is injective without truncation, but loses this injective property upon truncation. The second theorem says that, without truncation, PCA preserves pairwise distances, but upon truncation, PCA causes pairwise distances to either shrink or remain the same. In proving the theorems, I use facts from standard linear algebra texts, such as \cite{pool06} and \cite{stran03}. A statement of the first of the two theorems follows thus:

           \noindent \newtheorem{thm8}{Theorem }
          \begin{thm8}
             Suppose $C_x$ is any  real symmetric $n$ by $n$ matrix; suppose $Q^t$ is the $n$ by $n$ matrix formed by stacking the eigenvectors of $C_x$ one atop the other in increasing order of eigenvalues, from bottom up; and let $\hat{Q}^t$ be an $m$ by $n$ matrix, with $m < n$, obtained from $Q^t$ by discarding the $n - m$ lowermost rows of the latter. Then, for a given vector $\overline{\textbf{x}}$ $\in$ $\mathbb{R}^n$, and any vector $\textbf{x}$ $\in$ $\mathbb{R}^n$, the transformation  $\textbf{y(x)} = Q^t(\textbf{x} - \overline{\textbf{x}})$ is one-to-one, but the transformation $\hat{\textbf{y}}\textbf{(x)}  = \hat{Q}^t(\textbf{x} - \overline{\textbf{x}})$ is not. 
          \end{thm8}
           
            \begin{proof} [\textbf{Proof}]
          Denote by $\textbf{x}_i$ and $\textbf{x}_j$ any two distinct vectors in $\mathbb{R}^n$, so that one may write $\textbf{x}_i$ $\neq$ $\textbf{x}_j$, and one may set $\textbf{v} = \textbf{x}_j - \textbf{x}_i \neq 0$. Also, one may set  $\textbf{w} = \textbf{y}(\textbf{x}_j) - \textbf{y}(\textbf{x}_i) \;=\; Q^t(\textbf{x}_j - \overline{\textbf{x}}) - (Q^t(\textbf{x}_i - \overline{\textbf{x}}))  \;=\; Q^t(\textbf{x}_j - \textbf{x}_i)$. To show that $ \textbf{y}(\textbf{x}) = (Q^t(\textbf{x} - \overline{\textbf{x}}))$ is one-to-one, we must show that $\textbf{v} \neq \textbf{0} \: \Rightarrow \: \textbf{w} \neq \textbf{0}$. We therefore consider the linear system, $Q^t\textbf{v} = \textbf{w}$. When $\textbf{w} = \textbf{0}$, this linear system boils down to the homogeneous equation, $Q^t\textbf{v} = \textbf{0}$. Now, according to the fundamental theorem of linear algebra, if $Q^t$ is invertible, then the only solution to this homogeneous equation is the zero vector $\textbf{0}$. That is, up to $Q^t$ being invertible, we may write: $\textbf{w} = 0 \: \Rightarrow \: \textbf{v} = 0$, and the converse of this is: $\textbf{v} \neq 0 \: \Rightarrow \: \textbf{w} \neq 0$. Hence, what remains is to show that $Q^t$ is invertible. Firstly, by the fundamental theorem of linear algebra, $Q^{-1}$ exists. This is because the columns of $Q$ are the orthogonal, and therefore linearly independent, eigenvectors of the real symmetric matrix $C_x$. But, $(Q^t)^{-1} = (Q^{-1})^t$, from which we see that $Q^t$ is invertible. 
          
          The second part of the theorem asks us to prove that the transformation $\hat{\textbf{y}}\textbf{(x)}  = \hat{Q}^t(\textbf{x} - \overline{\textbf{x}})$ is not one-to-one. Towards this, we still set $\textbf{v} = \textbf{x}_j - \textbf{x}_i \neq 0$, as above, but now, we also set $\hat{\textbf{w}} = \hat{\textbf{y}}(\textbf{x}_j) - \hat{\textbf{y}}(\textbf{x}_i) = \hat{Q}^t(\textbf{x}_j - \textbf{x}_i)$. We need  to show that $\exists \textbf{v} \in \mathbb{R}^n$ such that $\textbf{v} \neq 0$, but $\hat{\textbf{w}} = 0$. To this end, we consider the  system $\hat{Q}^t\textbf{v} = \hat{\textbf{w}}$. When $\hat{\textbf{w}} = \textbf{0}$, this system reduces to the homogeneous equation, $\hat{Q}^t\textbf{v} = \hat{\textbf{0}}$,  which in augmented form is $[\hat{Q}^t|\textbf{0}]$. Now, matrix $\hat{Q}^t$ must be an $m$ by $n$ matrix with the property $m < n$, because $\hat{Q}$ results from discarding at least one row of an $n$ by $n$ matrix, $Q^t$. But, a theorem of linear algebra states that for any $m$ by  $n$ matrix $A$, if $m < n$ then the homogeneous system $[\hat{A}|\textbf{0}]$ has infinitely many solutions. Moreover, only one of these solutions is the zero vector, so that there exists infinitely many non-zero solutions. Thus, there is at least one non-zero solution in our homogeneous system, $\hat{Q}^t\textbf{v} = 0$. We conclude consequently that, in the system $\hat{Q}^t\textbf{v} = \hat{\textbf{w}}$, $\exists \textbf{v} \in \mathbb{R}^n$ such that $\textbf{v} \neq 0$, but $\hat{\textbf{w}} = 0$.   
           \end{proof}

          I make some physical comments about the just proven Theorem 1. Consider a classification problem with the  feature vectors denoted $\textbf{x}_i$. In particular, let us consider two \textbf{distinct} feature vectors $\textbf{x}_i$ and $\textbf{x}_j$, belonging to two different classes (or states of nature), $\omega_i$ and $\omega_j$.  According to Theorem 1, when the vectors $\textbf{x}_i$ and $\textbf{x}_j$ are passed through the truncated transformation of Equation 12, both vectors may end up as the same vector, $\textbf{y}_k$, on the output side of the transform. Definitely, this has the negative potential of causing the vectors to lose their discriminatory power as features for distinguishing between classes $\omega_i$ and $\omega_j$.
           
           To proceed, I state the second theorem of this manuscript. The theorem is in two parts, with the first part having already been referred to in the work of \cite{pcpre07}.
           
            \noindent \newtheorem{thm10} [thm8]{Theorem }
              \begin{thm10}
                Let $C_x$ be any  real symmetric $n$ by $n$ matrix; let $Q^t$ be the matrix whose rows are the eigenvectors of $C_x$; and let $\hat{Q}^t$ be a matrix obtained from $Q^t$ by discarding at least one row of the latter. Then, for a given vector $\overline{\textbf{x}}$ $\in$ $\mathbb{R}^n$, and any two  vectors $\textbf{x}_i$ $\in$ $\mathbb{R}^n$, and $\textbf{x}_j$ $\in$ $\mathbb{R}^n$, the pair of transformations  $\textbf{y}_i = \textbf{y}(\textbf{x}_i) = Q^t(\textbf{x}_i - \overline{\textbf{x}})$ and $\textbf{y}_j = \textbf{y}(\textbf{x}_j) = Q^t(\textbf{x}_j - \overline{\textbf{x}})$  satisfies  $|\textbf{x}_j - \textbf{x}_i| = |\textbf{y}_j - \textbf{y}_i|$, which preserves pairwise distances in $\mathbb{R}^n$; whereas the pair of transformations $\hat{\textbf{y}}_i = \hat{\textbf{y}}(\textbf{x}_i) = \hat{Q}^t(\textbf{x}_i - \overline{\textbf{x}})$ and $\hat{\textbf{y}}_j = \hat{\textbf{y}}(\textbf{x}_j) = \hat{Q}^t(\textbf{x}_j - \overline{\textbf{x}})$ satisfies the inequality $|\textbf{x}_j - \textbf{x}_i|   \geq |\hat{\textbf{y}}_j - \hat{\textbf{y}}_i|$, which tends to shrink pairwise distances in $\mathbb{R}^n$.  
              \end{thm10}

               \noindent \begin{proof} [\textbf{Proof}]
                      In what follows, we use $|.|$ to denote Euclidean distance.  So, for the first part of the theorem, we have $|\textbf{x}_j - \textbf{x}_i|^2 = (\textbf{x}_j - \textbf{x}_i)^t(\textbf{x}_j - \textbf{x}_i)$; and  $|\textbf{y}_j - \textbf{y}_i|^2 = (\textbf{y}_j - \textbf{y}_i)^t(\textbf{y}_j - \textbf{y}_i)$. But, $\textbf{y}_j - \textbf{y}_i = Q^t(\textbf{x}_j - \textbf{x}_i)$. Hence, $|\textbf{y}_j - \textbf{y}_i|^2 = (Q^t(\textbf{x}_j - \textbf{x}_i))^t Q^t(\textbf{x}_j - \textbf{x}_i) = (\textbf{x}_j - \textbf{x}_i))^tQQ^t(\textbf{x}_j - \textbf{x}_i) = (\textbf{x}_j - \textbf{x}_i))^t(\textbf{x}_j - \textbf{x}_i) = |\textbf{x}_j - \textbf{x}_i|^2$, where the penultimate equality is due to the orthogonality of $Q$. As a result, we easily find, $|\textbf{x}_j - \textbf{x}_i| = |\textbf{y}_j - \textbf{y}_i|$.  
                        
                        Now, for the second part of the theorem, given any vector $\textbf{x}_i$, we shall denote by $x_{ik}$ the $k$-th element of $\textbf{x}_i$. Further, we denote the $k$-th row of $Q^t$ by $(q^t)_k$ and the $k$-th row of $\hat{Q}^t$ by $(\hat{q}^t)_k$. We begin by considering the relationship $\textbf{y}_j - \textbf{y}_i = \hat{Q}^t(\textbf{x}_j - \textbf{x}_i)$, which derives easily from the hypothesis of the theorem. After using $(\hat{y}_j - \hat{y}_i)_k$ to denote the $k$-th element of the column vector $(\hat{\textbf{y}}_j - \hat{\textbf{y}}_i)_k$, one may write, $  ({y}_j - {y}_i)_k = (q^t)_k \: (\textbf{x}_j - \textbf{x}_i)$, from which one sees: $|\hat{\textbf{y}}_j - \hat{\textbf{y}}_i|^2 = \sum_{k = 1} ^{m} [(\hat{y}_j - \hat{y}_i)_k]^2 = \sum_{k = 1} ^{m} [\;(\hat{q}^t)_k \: (\textbf{x}_j - \textbf{x}_i)\;]^2$.  Now, on the other side, the relationship, $\textbf{y}_j - \textbf{y}_i = Q^t(\textbf{x}_j - \textbf{x}_i)$, allows us to write: $  ({y}_j - {y}_i)_k = (q^t)_k \: (\textbf{x}_j - \textbf{x}_i)$. Hence, $|\textbf{y}_j - \textbf{y}_i|^2 = \sum_{k = 1} ^{n} [(y_j - y_i)_k]^2 = \sum_{k = 1} ^{n} [\;(q^t)_k \: (\textbf{x}_j - \textbf{x}_i)\;]^2 = \sum_{k = 1} ^{m} [\;(q^t)_k \: (\textbf{x}_j - \textbf{x}_i)\;]^2  \; + \; \sum_{k = m + 1} ^{n} [\;(\hat{q}^t)_k \: (\textbf{x}_j - \textbf{x}_i)\;]^2$. Now since $\; \sum_{k = m + 1} ^{n} [\;(\hat{q}^t)_k \: (\textbf{x}_j - \textbf{x}_i)\;]^2 \geq 0$, it follows directly that $|\textbf{y}_j - \textbf{y}_i|^2 \geq |\hat{\textbf{y}}_j - \hat{\textbf{y}}_i|^2$. Consequently, we have: $|\textbf{y}_j - \textbf{y}_i| \geq |\hat{\textbf{y}}_j - \hat{\textbf{y}}_i|$. Now, in the first part of this theorem, we already established that $|\textbf{x}_j - \textbf{x}_i| = |\textbf{y}_j - \textbf{y}_i|$. Consequently, we see: $|\textbf{x}_j - \textbf{x}_i| = |\textbf{y}_j - \textbf{y}_i| \geq |\hat{\textbf{y}}_j - \hat{\textbf{y}}_i|$.     
                   \end{proof}

           \begin{figure} [h]
           \centering
           \includegraphics[scale = 1.0]{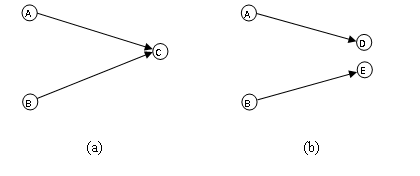} 
           \caption{ Illustration of how Theorem 2 is a ``milder" form of Theorem 1. 
            (a):  Depiction of the second part of Theorem 1. Points A and B are shrinked to a single point C by a truncated PCA transformation.(b): Depiction of the second part of Theorem 2. Points A and B are shrinked to two different points, D and E, by a truncated PCA transformation. The situation in (b) can be viewed as a ``milder" form of that in (a).}
           \end{figure}
           
           I use the just proven Theorem 2 to suggest how the error embodied in Equation 13 arises. That error is due to discarding the lowest $n - m$ rows of $Q^t$. More formally, notice that that error is a sum from $m + 1$ to $n$. Now, in the proof of Theorem 2, the``shrinking" actually occurs due to the term, $\; \sum_{k = m + 1} ^{n} [\;(\hat{q}^t)_k \: (\textbf{x}_j - \textbf{x}_i)\;]^2 $, which is again a sum from $m + 1$ to $n$. This leads me to opine that the truncation of $Q^t$ by $n - m$ rows leads to the shrinking effect, which in turn leads to the error underpinned by Equation 13. More so, there is another quite interesting connection between Theorems 1 and 2, in the sense that Theorem 2 can be viewed as a ``mild" form of Theorem 1. Theorem 1 says that truncated PCA is \textbf{not} one-to-one, so that it is possible for two points that were distinct before the transformation to``shrink"  to the same point after the transformation. On the other hand, Theorem 2 says that truncated PCA tends to cause points to shrink, although they may or may not shrink to the same point. Fig. 1 attempts to illustrate this connection.

\section{Connections between reconstruction errors and shrinkages in pair-wise distances}

In this section, we explore a connection between reconstruction errors engendered by the truncated PCA map and the associated shrinkages in pair-wise distances. We will show that, given any two points, $\textbf{x}_i$ and $\textbf{x}_j$  in $\mathbb{R}^n$, that are mapped by the truncated PCA transformation (or any other similar transformation for that matter) to the points $\hat{\textbf{y}}_i$ and $\hat{\textbf{y}}_j$ respectively; and if we define $e_{ij} = |\textbf{x}_i - \hat{\textbf{y}}_i | + | \textbf{x}_j - \hat{\textbf{y}}_j| $, as a form of total $L_1$ norm reconstruction error associated with $\textbf{x}_i$ and $\textbf{x}_j$ under the map; and also define $d_{ij} = |\textbf{x}_j - \textbf{x}_i| - |\hat{\textbf{y}}_j - \hat{\textbf{y}}_i| $ as the shrinkage in pair-wise distances associated with  $\textbf{x}_i$ and $\textbf{x}_j$ under the map, then we must have: $d_{ij} \: \le \: e_{ij}$. In words, this says that the pair-wise distance shrinkage associated with the pair of points, $\textbf{x}_i$ and $\textbf{x}_j$, is bounded from above by total $L_1$ norm reconstruction error associated with the pair of points. Indeed one can give a simple formal proof for the above statement, after stating it as a theorem:

\noindent \newtheorem{thm11} [thm8]{Theorem }
              \begin{thm11}
                Let  $\textbf{x}_i$ and $\textbf{x}_j$ be any two points in $\mathbb{R}^n$ which are mapped by the truncated PCA transformation to the points $\hat{\textbf{y}}_i$ and $\hat{\textbf{y}}_j$ respectively. Further, let $d_{ij} = |\textbf{x}_j - \textbf{x}_i| - |\hat{\textbf{y}}_j - \hat{\textbf{y}}_i|$ be the shrinkage in pairwise distances associated with the two points under the map; and let $r_{ij} = |\textbf{x}_i - \hat{\textbf{y}}_i | + | \textbf{x}_j - \hat{\textbf{y}}_j|$, be a total $L_1$ norm reconstrunction error associated with the two points under the map. Then, we must have: $ d_{ij} \le  r_{ij}$, which asserts that the shrinkage in pairwise distance is bounded from above by the total reconstruction error.
              \end{thm11}

\noindent \begin{proof} [\textbf{Proof}]
We begin with $d_{ij} = |\textbf{x}_j - \textbf{x}_i| - |\hat{\textbf{y}}_j - \hat{\textbf{y}}_i|$, and use the fact that, for any two complex numbers (or vectors), $z_1$ and $z_2$,  we must have $ |z_1 - z_2| \le |z_1| - |z_2|$ (which is a variant of the triangle inequality) to write  $|\textbf{x}_j - \textbf{x}_i| - |\hat{\textbf{y}}_j - \hat{\textbf{y}}_i| \le |\textbf{x}_j - \textbf{x}_i - \hat{\textbf{y}}_j + \hat{\textbf{y}}_i| = |(\textbf{x}_j - \hat{\textbf{y}}_j)  + (-\textbf{x}_i  + \hat{\textbf{y}}_i)|$. But, now by the fact $ |z_1 + z_2| \le |z_1| + |z_2|$, we have: $|(\textbf{x}_j - \hat{\textbf{y}}_j)  + (-\textbf{x}_i  + \hat{\textbf{y}}_i)| \le |\textbf{x}_j - \hat{\textbf{y}}_j| + | \hat{\textbf{y}}_i -\textbf{x}_i| = |\textbf{x}_j - \hat{\textbf{y}}_j| +|\textbf{x}_i - \hat{\textbf{y}}_i | = r_{ij}$. Hence, $ d_{ij} \le  r_{ij}$ as required.
 
\end{proof}

To re-iterate, the just proven theorem shows a connection between an $L_1$ norm reconstruction error and pair-wise distance shrinkages. But, intuitively, one feels that this $L_1$ norm reconstruction error should be correlated with the root mean square reconstruction error. Now, according to Equation 13, the root mean square error is simply the sum of those eigenvalues associated with the discarded eigenvector rows in the truncated PCA map. Hence, I was lead to perform numerical experiments aimed at calculating the correlation between the sum of those eigenvalues and shrinkages in pair-wise distances. In addition, I have also performed some experiments to check respectively the effect of the sum of those eigenvalues and the  effect of the shrinkages on classification accuracies under the PCA map. So far, I have obtained the following results on some publicly available data from the UCI Machine Learning Repository:  1). There seems to be a strong correlation between the sum of those eigenvalues associated with discarded eigenvectors and shrinkages in pair-wise distances. 2). Neither the sum of those eigenvalues nor pair-wise distances have any strong correlations with classification accuracies.

\section{Conclusion}
This \lq on-going\rq $\:$ work presented some theoretical and empirical implications of data transformations under the PCA.  In particular,  following three results about PCA maps were stated and proven:  1). PCA without discarding eigenvector rows is injective, but looses this injectivity when eigenvector rows are discarded  2). PCA without discarding eigenvector rows preserves pair-wise distances, but tends to cause pair-wise distances to shrink when eigenvector rows are discarded. 3). For any pair of points, the shrinkage in pair-wise distance is bounded above by an $L_1$ norm reconstruction error associated with the points.  Further, since the third result  suggests that there might exist  some correlation  between shrinkages in pair-wise distances and mean square reconstruction error which is defined as the sum of those eigenvalues associated with the discarded eigenvectors, I was naturally led to perform numerical experiments to obtain the correlation between the sum of those eigenvalues and shrinkages in pair-wise distances. In addition, I have also performed some experiments to check respectively the effect of the sum of those eigenvalues and the  effect of the shrinkages on classification accuracies under the PCA map. So far, the following results have been obtained on some publicly available data from the UCI Machine Learning Repository:  1). There seems to be a strong correlation between the sum of those eigenvalues associated with discarded eigenvectors and shrinkages in pair-wise distances. 2). Neither the sum of those eigenvalues nor pair-wise distances have any strong correlations with classification accuracies.

\end{document}